\documentclass[11pt,a4paper]{article}
\usepackage[hyperref]{acl2020}
\usepackage{times}
\usepackage{latexsym}
\hypersetup{breaklinks=true}
\usepackage{microtype}

\usepackage{amsmath,amsfonts,bm}

\def\eqref#1{equation~\ref{#1}}
\def\1{\bm{1}}

\DeclareMathAlphabet{\mathsfit}{\encodingdefault}{\sfdefault}{m}{sl}
\SetMathAlphabet{\mathsfit}{bold}{\encodingdefault}{\sfdefault}{bx}{n}

\usepackage{url}
\usepackage{times}
\usepackage{latexsym}
\usepackage{amssymb}
\usepackage{soul}
\usepackage{textcomp}
\usepackage{gensymb}
\usepackage{pgfplots}
\usepackage{url}
\usepackage[utf8]{inputenc}
\usepackage{comment}
\usepackage{array}

\usepackage{listings}
\usepackage{setspace}
\newcolumntype{L}{>{\arraybackslash}m{15.0cm}}

\usepgfplotslibrary{external}
\pgfplotsset{width=10cm,compat=1.9}

\newcommand{\lv}[1]{}

\title{Joint translation and unit conversion for end-to-end localization}

\author{Georgiana Dinu  \hspace{0.075cm}
  Prashant Mathur \hspace{0.075cm}
  Marcello Federico \hspace{0.075cm}
  Stanislas Lauly \hspace{0.075cm}
  Yaser Al-Onaizan \\\\
  Amazon AWS AI\\
  \small{\texttt{\{gddinu,pramathu,marcfede,laulysl,onaizan\}@amazon.com}}
  }
\aclfinalcopy
\begin{document}

\maketitle

\begin{abstract}
A variety of natural language tasks require processing of textual data which contains a mix of natural language and formal languages such as mathematical expressions. In this paper, we take unit conversions as an example and propose a data augmentation technique which lead to models learning both translation and conversion tasks as well as how to adequately switch between them for end-to-end localization.
\end{abstract}

%introduction.tex
\section{Introduction}

Neural networks trained on large amounts of data have been shown to achieve state-of-the art solutions on most NLP tasks such as textual entailment, question answering, translation, etc. In particular, these solutions show that one can successfully model the ambiguity of language by making very few assumptions about its structure and by avoiding any formalization of language. However, unambiguous, formal languages such as numbers, mathematical expressions or even programming languages (e.g. markup) are abundant in text and require the ability to model the symbolic, ``procedural" behaviour governing them. \citep{DBLP:journals/corr/abs-1901-03735,dua-etal-2019-drop}.

An example of an application where such examples are frequent is the extension of machine translation to localization. Localization is the task of combining translation with ``culture adaptation", which involves, for instance, adapting dates (\textit{12/21/2004} to \textit{21.12.2004)}, calendar conversions (\textit{March 30, 2019} to \textit{Rajab 23, 1441} in Hijri Calendar) or conversions of currencies or of units of measure (\textit{10 kgs} to \textit{22 pounds}).

Current approaches in machine translation handle the processing of such sub-languages in one of two ways: The sub-language does not receive any special treatment but it may be learned jointly with the main task if it is represented enough in the data. Alternatively, the sub-language is decoupled from the natural text through pre/post processing techniques: e.g. a \textit{miles} expression is converted into \textit{kilometers} in a separate step after translation.

Arguably the first approach can successfully deal with some of these phenomena: e.g. a neural network may learn to invoke a simple conversion rule for dates, if enough examples are seen training.
However, at the other end of the spectrum, correctly converting distance units, which itself is a simple algorithm, requires knowledge of numbers, basic arithmetic and  the specific conversion function to apply. It is unrealistic to assume a model could learn such conversions from limited amounts of parallel running text alone.
Furthermore, this is an unrealistic task even for distributional, unsupervised pre-training \cite{Turney:2010,baroni:2010,peters-etal-2018-deep}, despite the success of such methods in capturing other non-linguistic phenomena such as world knowledge or cultural biases \cite{Bolukbasi:2016,Van:2018}.\footnote{\citep{wallace-etal-2019-nlp} show that numeracy is encoded in pre-trained embeddings. While promising, this does not show that more complex and varied manipulation of numerical expressions can be learned in a solely unsupervised fashion.}

While the second approach is currently the preferred one in translation technology, such decoupling methods do not bring us closer to end-to-end solutions and they ignore the often tight interplay of the two types of language: taking unit conversion as an example, \textit{approximately 500 miles}, should be translated into \textit{ungef\"ahr 800 km} (approx. 800km) and not \textit{ungef\"ahr 804 km} (approx. 804km).

In this paper we highlight several of such language mixing phenomena related to the task of localization for translation and focus on two distance (miles to kilometers) and temperature (Fahrenheit to Celsius) conversion tasks. Specifically, we perform experiments using the popular MT transformer architecture and show that the model is successful at learning these functions from symbolically represented examples. Furthermore, we show that data augmentation techniques together with small changes in the input representation produce models which can both translate and appropriately convert units of measure in context.

\section{Related work}

Several theoretical and empirical works have addressed the computational capabilities end expressiveness of deep learning models. Theoretical studies on language modeling  have mostly targeted simple grammars from the Chomsky hierarchy.
In particular, \citet{hahn_2019} proves that Transformer networks suffer limitations in modeling  regular periodic  languages (such as $a^n b^n$)
as well as  hierarchical (context-free) structures, unless their depth or self-attention heads increase with the input length. On the other hand,
\citet{merrill2019sequential} proves that LSTM  networks can recognize a subset of periodic languages. Also experimental papers  analyzed the capability
of LSTMs to recognize these two language classes \cite{weiss_practical_2018,suzgun_evaluating_2019,sennhauser_evaluating_2018,skachkova_closing_2018, bernardy_can_2018},  as well as natural language hierarchical structures  \cite{linzen_2016,gulordava_colorless_2018}.
\lv{In particular, \citet{tran_2018}  showed that LSTM  outperform Transformer models in learning  English dependency structures. }
\lv{To overcome these shortcoming, Universal Transformers \cite{DBLP:conf/iclr/DehghaniGVUK19} were recently
proposed, which integrate self-attention with a layer-recurrent (not time-recurrent)  topology.}  It is worth noticing, however,  that differently
from formal language recognition tasks,  state of the art machine translation systems \cite{wmt-2019-findings,IWSLT-2019-findings} are still based on the Transformer architecture .

\lv{\citet{lin_2017}
shows that feed forward networks can  approximate low-order polynomial Hamiltonians, which model well properties of physical-world data,
such as symmetry, locality and compositionality. Classic  \cite{siegelmann_computational_1995} and recent \cite{perez_turing_2019} results show,
respectively,  that recurrent networks and Transformer networks can emulate Turing machines, assuming unbounded computations
on their input.}

Other related work addresses specialized neural architectures capable to process and reason with numerical expressions for binary addition, evaluating arithmetic expressions or other number manipulation tasks \citep{Joulin:2015:IAP:2969239.2969261, SaxtonMath, NIPS2018_8027, DBLP:journals/corr/abs-1809-08590}. While this line of work is very relevant, we focus on the natural intersection of formal and everyday language. The types of generalization that these studies address, such as testing with numbers orders of magnitude larger than those in seen in training, are less relevant to our task.

The task of solving verbal math problems \cite{mitra-baral-2016-learning, wang-etal-2017-deep, koncel-kedziorski-etal-2016-mawps,SaxtonMath} specifically addresses natural language mixed with formal language. Similarly, \citep{DBLP:journals/corr/abs-1901-03735} introduces a benchmark for evaluating quantitative reasoning in natural language inference and \citep{dua-etal-2019-drop} one for symbolic operations such as addition or sorting in reading comprehension. However these papers show the best results with two-step approaches, which extract the mathematical or symbolic information from the text and further manipulate it analytically. We are not aware of any other work successfully addressing both machine translation and mathematical problems, or any of the benchmarks above, in an end-to-end fashion.

\section{Unit conversion in MT localization}
The goal of localization is to enhance plain content translation so that the final result looks and feels as being created for a specific target audience. \lv{It thus involves combining translation with adopting local formats (dates, times, addresses and phone numbers), currencies and units of measure, up to adapting names, idioms, colors and graphics to the local culture and taste.}

Parallel corpora in general include localization of formats %, such as dates (e.g. \emph{11/30/2019} (en-us) $\rightarrow$ \emph{30/11/2019} (en-uk)) and
numeric expressions (e.g. from \emph{1,000,000.00} (en-us) to \emph{1.000.000,00} (de-de)). Format conversions in most of the cases reduce to operations such as reordering of elements and replacement of symbols, which quite naturally fit inside the general task of machine translation. In this paper, we are interested in evaluating the capability of neural MT models to learn less natural operations, which are typically involved in the conversion of time expressions (e.g. 3:30pm $\rightarrow$ 15:30) and units of measure, such as lengths (10ft to 3m) and temperatures (55F to 12.8C).
\lv{While such conversions can be addressed via post-processing and simple calculations, our investigation aims to see to what extend a neural model can jointly learn and combine translation and computation skills, such as simple linear transformations.}

We choose two measure unit conversion tasks that are very prevalent in localization: Fahrenheit to Celsius temperature conversion and miles to kilometers. We address the following questions: 1) Can a standard NMT architecture, the transformer, be used to learn the functions associated with these two conversion tasks (Section 3.1) and 2) Can the same architecture be used to train a model that can do both MT and unit conversion? (Section 3.2)

\subsection{Unit conversion}

\paragraph{Network architecture}
We use the state-of-the-art transformer architecture \cite{DBLP:conf/nips/VaswaniSPUJGKP17} and the Sockeye Toolkit \cite{DBLP:journals/corr/abs-1712-05690} to train a network with 4 encoder layers and 2 decoder layers for a maximum of 3000 epochs (See Appendix A for details). As the vocabulary size is small the training is still very efficient. For the experiments training several tasks jointly we facilitate the context-switching between the different tasks with an additional token-level parallel stream (source factors) \citep{sennrich-haddow-2016-linguistic}. We use two values for the digits in numerical expressions (distance/temperature) and a third value for all other tokens. These are concatenated to each token as 8-dimensional embeddings.
%The input representation consists of 32,000 BPE units~\cite{Sennrich15}, shared across the source and target language, and all numbers associated with the units to be converted are split into single digit characters.
\paragraph{Data}
The models are trained with parallel examples of the two functions, one affine: $\mathrm{\degree F \rightarrow {\degree}C}(x) = (x - 32) \times \frac{5}{9}$ and one linear: $\mathrm{mi \rightarrow km}(x) = x \times 1.60934$.
For each task, we generate training data of various input lengths ranging from 1 to 6 digits in the input. The input is distributed uniformly w.r.t 1) integer versus single digit precision (with the output truncated to same precision as the input) and 2) the length of the input in digits. We over-sample when there are not enough distinct data points, such as in the case of double- or single-digit numbers. The numerical input is tokenized into digits (e.g. \textit{5 2 1 miles}) and we train individual models for the two functions, as well as joint models, using held-out data for validation and testing. Note that unlike previous work, we are interested only in interpolation generalization: test numbers are unseen, but the \textit{range} of test numbers does not increase.

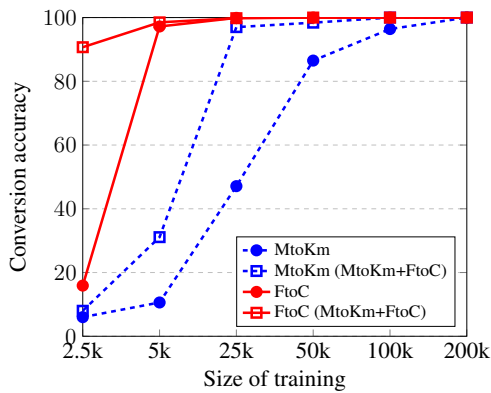
\begin{figure}
\centering
\begin{tikzpicture}[scale=0.6]
\begin{axis}[
    xlabel={Size of training},
    ylabel={Conversion accuracy},
    xmin=0, xmax=100,
    ymin=0, ymax=100,
    xtick={0,20,40,60,80,100},
    xticklabels={2.5k,5k,25k,50k,100k,200k},
    ytick={0,20,40,60,80,100},
    legend pos=south east,
    legend cell align={left},
    ymajorgrids=true,
    grid style=dashed,
    mark size=3.0pt,
    label style={font=\Large},
    tick label style={font=\Large}
]
 \addplot[
    color=blue,
    mark=*,
    dash pattern=on 3pt off 3pt,
    mark options={solid, fill=blue},
    ][ultra thick]
    coordinates {
    (0,6.0)(20,10.6)(40,47.1)(60,86.5)(80,96.4)(100,100)
    };
 \addplot[
    color=blue,
    mark=square,
    dash pattern=on 3pt off 3pt,
    mark options={solid, fill=blue},
    ][ultra thick]
    coordinates {
    (0,8.0)(20,31.1)(40,97.0)(60,98.4)(80,100)(100,100)
    };
 \addplot[
    color=red,
    mark=*,
    mark options={solid, fill=red},
    ][ultra thick]
    coordinates {
    (0,15.9)(20,97.2)(40,99.9)(60,100)(80,100)(100,100)
    };
 \addplot[
    color=red,
    mark=square,
    mark options={solid, fill=red},
    ][ultra thick]
    coordinates {
    (0,90.7)(20,98.5)(40,99.8)(60,100)(80,100)(100,100)
    };
    \legend{MtoKm,MtoKm (MtoKm+FtoC), FtoC, FtoC (MtoKm+FtoC)}
 \end{axis}
\end{tikzpicture}
\caption{\footnotesize{Conversion accuracy with $\pm10^{-4}$ tolerance on relative error, as a function of the number of the target conversion examples in the train data. Functions are learned both in isolation and in a joint setting (MtoKm + FtoC) which adds to training an equal amount of data for the other function.}}
 \label{figure:results 1}
\vspace{-0.5cm}
\end{figure}

\paragraph{Results}
Results as a function of the amount of training data are given in Figure \ref{figure:results 1}. Test sets are synthetic and contain numbers in $\mathrm{[10^3-10^6]}$ range.

The results show that the transformer architecture can learn the two functions perfectly, however, interestingly enough, the two functions are learned differently. While the degree conversion is learned with a high accuracy with as little as several thousand examples, the distance conversion is learned gradually, with more data leading to better and better numerical approximations: in this case the model reaches high precision in conversion only with data two orders of magnitude larger. Both functions are learned with less data when training is done jointly and source factors are used - this suggests that, despite the fact that the functions are very different, joint training may facilitate the learning of numbers as a general concept and helps learn additional functions more efficiently.
\lv{Another observation is that while the model learns large number conversion perfectly when given sufficient data, it has difficulties generalizing from large numbers to smaller numbers ($<10^3$), as the number of distinct examples that can be provided in the training is small. Previous literature has investigated generalizing to increasingly larger input strings, however the reversed generalization seems to be difficult as well, although this specific case can be practically solved by all these examples in the training. These experiments show that while conversion can be learned perfectly, the mechanisms through which the neural networks learn this are still largely unknown.}

\subsection{Joint MT and unit conversion}

\begin{table*}[ht]
    \centering
    {\small{
    \begin{tabular}{c|l|l}
      & \multicolumn{2}{l}{Example} \\
    \hline
    {Conv} & 5 2 1 miles & 8 3 9 km \\
    \hline
    MT & We do not know what is happening. & Wir wissen nicht, was passiert.\\
    \hline
    Loc. & The venue is within 3 . 8 miles from the city center& Die Unterkunft ist 6 km vom Stadtzentrum entfernt\\
    \hline
    \end{tabular}
    \caption{\footnotesize{The three types of data used in training the joint model: unit conversion data, standard MT data and localization (Loc) data containing unit conversions in context.}}
    \label{tab:examples}
    }}
\end{table*}

In a second set of experiments we investigate if the transformer model is able to perform both the translation and the unit conversion tasks and learns to adequately switch from one to the other in context. We use the same architecture as in the previous section, with minor modifications: we use subword embeddings with a shared vocabulary of size 32000 and a maximum number of epochs of 30.
\begin{table}[ht]
%\vspace{0.1cm}
    \centering
    {\small{
    \begin{tabular}{cc|ccccc}
      &  & news17 & \multicolumn{2}{c}{Loc-dist} & \multicolumn{2}{c}{Loc-temp} \\
      S.f. & \#Loc & Bleu & Bleu & Acc. & Bleu & Acc.\\
    \hline
    - & 0     &  22.7 & 20.6 & 0\%   & 16.1 &  0\% \\
    - & 5k    &  22.7 & 56.7 & 52.3\%& 44.1 & 48.3\% \\
    - & 15k   &  23.0 &  61.7  & 76.2\% & 48.5& 80.3\% \\
    - & 30k   &  23.0 &  65.0  & 90.3\% & 48.9& 81.3\% \\
    \hline
    \checkmark & 0     &  22.9 & 19.5 & 1\%   & 16.6 & 3.4\% \\
    \checkmark & 5k    &  22.9 & 58.7 & 69.4\%& 46.8 & 64.8\% \\
    \checkmark & 15k   &  23.2 &  63.0  & 88.0\% & 48.6& 77.8\% \\
    \checkmark & 30k   &  22.6 &  64.0  & 88.3\% & 48.8& 79.4\% \\
    \end{tabular}
    \caption{\footnotesize{Bleu scores and accuracy on conversion of degrees (temp) and miles (dist) expressions in Loc test sets. Conversion accuracy is computed with a tolerance of 0.01\%. All models are trained using: 2.2M MT+ 100k Conv + \#Loc data (col 2) for each function, with and without \textbf{S}ource \textbf{f}actors (column 1).}}
    \label{tab:results 3}
    }}
    \vspace{-0.2cm}
\end{table}

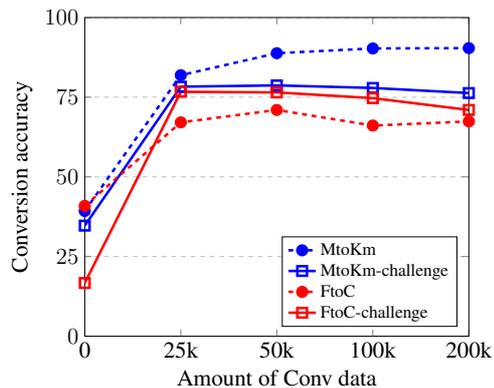
\begin{figure}
\centering
\begin{tikzpicture}[scale=0.6]
\begin{axis}[
    xlabel={Amount of Conv data},
    ylabel={Conversion accuracy},
    xmin=0, xmax=100,
    ymin=0, ymax=100,
    xtick={0,25,50,75,100},
    xticklabels={0,25k,50k,100k,200k},
    ytick={0,25,50,75,100},
    legend pos=south east,
    legend cell align={left},
    ymajorgrids=true,
    grid style=dashed,
    mark size=3.0pt,
    label style={font=\Large},
    tick label style={font=\Large}
]
 \addplot[
    color=blue,
    mark=*,
    dash pattern=on 3pt off 3pt,
    mark options={solid},
    ][ultra thick]
    coordinates {
    (0,39.3)(25,81.9)(50,88.8)(75,90.3)(100,90.4)
    };
 \addplot[
    color=blue,
    mark=square,
    ][ultra thick]
    coordinates {
    (0,34.7)(25,78.3)(50,78.7)(75,77.9)(100,76.3)
    };
 \addplot[
    color=red,
    mark=*,
    dash pattern=on 3pt off 3pt,
    mark options={solid},
    ][ultra thick]
    coordinates {
    (0,40.9)(25,67.1)(50,71.0)(75,66.1)(100,67.4)
    };
 \addplot[
    color=red,
    mark=square,
    ][ultra thick]
    coordinates {
    (0,16.7)(25,76.7)(50,76.5)(75,74.7)(100,71.0)
    };
    \legend{MtoKm,MtoKm-challenge,FtoC,FtoC-challenge}
 \end{axis}
\end{tikzpicture}
\caption{\footnotesize{Accuracy of localization conversion (tolerance 0.01\%) on regular and \textit{challenge} sets. All models use source factors and are trained using: 2.2M MT data + 15k Loc data + varying amounts of Conv data.}}
 \label{figure:results 2}
 \vspace{-0.3cm}
\end{figure}

\paragraph{Data}
As standard MT parallel data we use a collection containing Europarl~\cite{koehn2005epc} and news commentary data from WMT En$\to$De shared task 2019 totalling 2.2 million sentences.\footnote{We opt for a smaller experiment in order to speed up computations and to prioritize efficiency in our experiments \citep{strubell-etal-2019-energy}. We have no reason to assume any dependency on the data size.} Standard translation test sets do not have, however, enough examples of unit conversions and in fact corpora such as CommonCrawl show inconsistent treatment of units. For this reason, we create a unit conversion (Localization) data set. %We employ two simple strategies to train a single model to do both translation and conversion tasks, data augmentation  and enhanced input representation through source factors.
We extract sentences containing Fahrenheit/Celsius and miles/km from a mix of open source data sets namely, ParaCrawl, DGT (Translation Memories), Wikipedia and OpenSubtitles, TED talks from OPUS \cite{tiedemann12}. Regular expressions are used to extract the sentences containing the units and modify the source or the reference by converting the matched units. For example, if \textit{5 km} is matched in the reference, we modify the source expression to \textit{3.1 miles}.\footnote{Scripts to create this data will be released, however the data used itself does not grant us re-distribution rights.} We are able to extract a total of 7k examples for each of the two conversion tasks and use 5k for training and 2k for testing, making sure the train/test numerical expressions are distinct. \lv{The total number of distinct conversions found in data is 1.4k for distance and 0.4k for temperature. We split these evenly between train and test, making sure the two conversion sets, as well as the sentences containing them, do not overlap.
While small, the amount of localization data we use in training matches real-world scenarios as it is unrealistic to assume that such data can be easily sourced for the potentially diverse set of conversions that may appear in various domains. }

\paragraph{Results} In the experimental setting, we distinguish the following three types of data: translation (\textbf{MT}), conversion (\textbf{Conv}) and localization data (conversion in context) (\textbf{Loc}), and measure performance when varying amounts of Conv and Loc are used in training. Examples of these data types are given in Table \ref{tab:examples}.
%All the models achieve 100\% accuracy at conversion in isolation, which we omit in the table for brevity.
The first set of experiments (Table \ref{tab:results 3}) uses MT and Conv data and tests the models' performance with varying amounts of Loc data. We observe that for localization performance, Loc data in training is crucial: accuracy jumps from 2\% when no Loc data is used to 66\% for 5k Loc and to 82\%, on average, with 15k localization examples for each function (w. source factors). However, the 15k data points are obtained by up-sampling the linguistic context and replacing the unit conversions with new unit conversions, and therefore no ``real" new data is added. We observe no further improvements when more Loc data is added. Regarding the use of source factors, they help when the localization data is non-existent or very limited, however their benefits are smaller otherwise.

The Bleu scores measured on a news data set as well as on the localization data sets show no degradation from a baseline setting, indicating that the additional data does not affect translation quality. The exception is the \#Loc-0 setting, in which the model wrongly learns to end all localization sentences with \textit{km} and \textit{C} tokens respectively, as seen in the Conv data. Similarly to the previous results, temp conversions are learned either correctly or not at all while the distance ones show numerical approximation errors: When measuring exact match in conversion (0.0 tolerance), the temperature accuracy remains largely the same while the distance accuracy drops by up to 30\%.

Given the observation that Loc data is crucial, we perform another set of experiments to investigate if the Conv data is needed at all. Results are shown in Figure \ref{figure:results 2}. In light of the limited amount of real distinct conversions that we see in testing, we create two additional challenge sets which use the same linguistic data and replace the original conversions with additional ones uniformly distributed w.r.t the length in digits from 1 to 6. The results indicate that conversion data is equally critical, and that the conversion cannot be learned from the localization data provided alone. The localization data rather acts as a ``bridge" allowing the network to combine the two tasks it has learned independently.

\section{Conclusions}
We have outlined natural/formal language mixing phenomena in the context of end-to-end localization for MT and have proposed a data augmentation method for learning unit conversions in context.
Surprisingly, the results show not only that a single architecture can learn both translation and unit conversions, but can also appropriately switch between them when a small amount of localization data is used in training. For future work we plan to create a diverse localization test suite and investigate if implicit learning of low-level concepts such as natural numbers takes place and if unsupervised pre-training facilitates such learning.

%\bibliography{acl2020}
%\bibliographystyle{acl_natbib}

\newpage
\appendix
\section{Appendix}

\begin{lstlisting}
encoder-config:
  act_type: relu
  attention_heads: 8
  conv_config: null
  dropout_act: 0.1
  dropout_attention: 0.1
  dropout_prepost: 0.1
  dtype: float32
  feed_forward_num_hidden: 2048
  lhuc: false
  max_seq_len_source: 101
  max_seq_len_target: 101
  model_size: 512
  num_layers: 4
  positional_embedding_type: fixed
  postprocess_sequence: dr
  preprocess_sequence: n
  use_lhuc: false

decoder config:
  act_type: relu
  attention_heads: 8
  conv_config: null
  dropout_act: 0.1
  dropout_attention: 0.1
  dropout_prepost: 0.1
  dtype: float32
  feed_forward_num_hidden: 2048
  max_seq_len_source: 101
  max_seq_len_target: 101
  model_size: 512
  num_layers: 2
  positional_embedding_type: fixed
  postprocess_sequence: dr
  preprocess_sequence: n

config_loss: !LossConfig
  label_smoothing: 0.1
  name: cross-entropy
  normalization_type: valid
  vocab_size: 32302

config_embed_source: !EmbeddingConfig
  dropout: 0.0
  dtype: float32
  factor_configs: null
  num_embed: 512
  num_factors: 1
  vocab_size: 32302

config_embed_target: !EmbeddingConfig
  dropout: 0.0
  dtype: float32
  factor_configs: null
  num_embed: 512
  num_factors: 1
  vocab_size: 32302
\end{lstlisting}

\end{document}